# Context-LSTM: a robust classifier for video detection on UCF101

Dengshan Li[1,2,3], Rujing Wang[1,3]

[1]Institute of Intelligent Machines, Hefei Institutes of Physical Science, Chinese Academy of Sciences,

[2]Science Island Branch of Graduate School, University of Science and Technology of China,

[3]Intelligent Agriculture Engineering Laboratory of Anhui Province

## Abstract

Video detection and human action recognition may be computationally expensive, and need a long time to train models. In this paper, we were intended to reduce the training time and the GPU memory usage of video detection, and achieved a competitive detection accuracy. Other research works such as Two-stream, C3D, TSN have shown excellent performance on UCF101. Here, we used a LSTM structure simply for video detection. We used a simple structure to perform a competitive top-1 accuracy on the entire validation dataset of UCF101. The LSTM structure is named Context-LSTM, since it may process the deep temporal features. The Context-LSTM may simulate the human recognition system. We cascaded the LSTM blocks in PyTorch and connected the cell state flow and hidden output flow. At the connection of the blocks, we used ReLU, Batch Normalization, and MaxPooling functions. The Context-LSTM could reduce the training time and the GPU memory usage, while keeping a state-of-the-art top-1 accuracy on UCF101 entire validation dataset, show a robust performance on video action detection.

## 1. Introduction

Machine learning has developed rapidly. Machine learning is a combination of a series of non-linear function layers, which is used for computer vision, pattern recognition, audio processing, natural language processing, etc. With a series of successes, deep learning has become one of the main research directions.

Deep neural networks have been used in image and video detection, and have realized good results. AlexNet [1] greatly improves the accuracy of image recognition. VGG net [2] improves further. The residual shortcut connection in ResNet [3] reduces the loss of information transmitted by the network, and increases the depth of the network layer, and greatly improves the detection accuracy. GoogLeNet [4-6] expands the depth and width of the network, making the features extracted by the network have a higher dimension. RNN [7] is proposed for the learning of temporal series. LSTM [8] is a change of RNN, which is used to prevent the gradient explosion or disappearance in the learning of temporal series.

The proposal of LSTM is a great progress for the temporal series learning, since the speed and accuracy of temporal series detection are both enhanced. LSTM uses the temporal information between the temporal series, and can better learn the features with the temporal information. In the traditional LSTM learning architecture, the feature extraction of the temporal series often passes through multi-layer neural networks, and these multi-layer neural networks may lose part of the data feature information. We think that, using ResNet's shortcut connection to these multi-layer neural networks, or using bi-directional LSTM (RNN) will reduce the loss of the data feature information.

In this study, we propose a kind of LSTM structure, which is named Context-LSTM, for the purpose of using the context information of the temporal sequences more. This structure adopts the way of hierarchical cascade of LSTM, and connects the cell data stream and hidden data stream of LSTM end-to-end in each layer. We believe that the structure can enhance the utilization of the temporal information of the input data.

The main contributions of the paper are as follows:

a) The information transmission loss during model training is proposed and analyzed.

b) The deep temporal feature is defined as the feature extracted by a deep neural network from a temporal sequence.

c) The Context-LSTM classifier is proposed to process the deep temporal feature, which reached a state-of-the-art top-1 accuracy on the entire validation dataset of UCF101. The proposed classifier is robust, stable and reliable.

d) The two-stage of human recognition action system is simulated by the proposed Context-LSTM system.

## 2. Related work

Shi et al. [9] applied LSTM method to precipitation nowcasting, and outperformed the state-of-the-art algorithms for precipitation nowcasting. They proposed ConvLSTM which includes the encoding network and the forecasting network, for making the predict. Duan et al. [10] introduced a novel framework named OmniSource to train action recognition models. The framework used a teacher network and a student network for training, and the framework realized state-of-the-art detection accuracy on several datasets. Wang et al. [11] introduced I3D-LSTM, which used the 3D ConvNet with Inception module as the backbone, LSTM as the classifier. I3D-LSTM achieved state-of-the-art detection accuracy on UCF101. Huang et al. [12] defined Busy-Quiet Net (BQN), which could distill the redundancy of adjacent frames. Gowda et al. [13] proposed a frame selection method for the video classification, the method could select the frames with more temporal and object feature information. Shalmani et al. [14] proposed a confidence distillation inference framework for action recognition. The framework used a teacher network and a student network to train, and the teacher network and the student network used different frames of an entire video. Zhou et al. [15] proposed C-LSTM for classifying the text, which incorporated a convolutional neural network into the cells of an LSTM. C-LSTM reached excellent performance on text classification.

## 3. Materials and methods

LSTM is suitable for the feature extraction of temporal series, since the structure of LSTM has the temporal features. Our study is to enhance the utilization of these temporal information by LSTM.

With more layers of neural network, the semantic level of the extracted features could become higher. However, with more network layers, the loss of data feature information may also be more. The shortcut of ResNet could solve the problem of the information loss when the network is deeper, and we use ResNet-50 as the backbone (feature extractor).

The architecture we proposed includes a ResNet-50 backbone and the LSTM module. The LSTM module is used as the classifier. The LSTM module makes use of the temporal information among the input data well. The module adopts a cascade structure to connect the LSTM sequences step by step. Among the LSTM blocks, we use the ReLU activation function and the Batch Normal [16] function for the connection. We interconnect the cell state flow and hidden output flow of LSTM at each block, in order to increase the temporal characteristics of the LSTM network structure.

### 3.1 Dataset

The dataset we used is UCF101 [17], which is a human action video dataset. The UCF101 dataset contains much temporal information since the video frames have many correlations.

The UCF101 dataset contains 101 human action categories. It consists of 13320 small videos. The duration of each video is from a few seconds to more than ten seconds. The resolution of the video frames is $320\times240$. There is only one category in one video. The total duration length of the 13320 videos is nearly 27 hours.

### 3.2 Deep temporal feature

The development trend of artificial neural networks is to increase the network layers, and extract more advanced semantic features. One of the classic examples is ResNet. Before ResNet, the increasing of the network layers increased the information loss through the information transmission. As a result, the deepening of the network might reduce the detection accuracy. However, the shortcut connection in ResNet reduces the information loss, and a deeper network could extract more advanced semantic features, and then could greatly improve the detection accuracy.

Deeper networks should extract more advanced semantic features, and the detection effect can be improved. We apply this method to the LSTM system. By using the deeper neural network, the LSTM system can extract more advanced semantic features. We use a deeper LSTM network layers, for the purpose of extracting higher level features. The experiments show that, our proposed structure reduces the model training time and improves the detection accuracy with the same epochs. Moreover, our model occupies less GPU memory than the ordinary LSTM structure (ResNet-50-LSTM), reducing an average of nearly 2GB GPU memory in our experiments.

The features extracted by the deep convolutional neural network are then sent into the deep temporal feature extractor and classifier (such as LSTM in our experiments), which is named deep temporal features. Deep temporal features have higher semantic features, which can transfer deeper semantic temporal features among the temporal feature classifiers. Thus, it can enhance the detection of the temporal sequences.

We define temporal information as the correlation among the temporal sequences. Using temporal information is a good way of detecting the temporal sequences, such as video frames, language words, etc. The deep temporal features can be expressed by LSTM, FlowNet [18], etc., which is more helpful for the long sequence detection.

## 3.3 The information loss during the neural network transmission

We assume that, during the transmission process of artificial convolutional neural network, the information contained in the input data may have some loss. For instance, the convolution operation is to multiply the corresponding elements of the feature map and the convolution kernel in adjacent regions, and then sum every product. Therefore, the multiple elements which are the size of the convolution kernel are transformed into one element. This compression may be the possible information loss.

Similarly, the pooling operation is to transform the elements of adjacent regions into one element, by computing the maximum value or the average value, etc. This is also a kind of compression essentially. The compression may cause the information loss either. The ReLU operation filters out the elements which are less than zero, so that the information contained in the element value may have the loss.

Based on the above analysis, we think that the information should have some loss in the network transmission. And then, how to reduce this loss might be a direction for the neural network study. The residual shortcut connection of ResNet is one of the ways to reduce the information loss, because the shortcut connection directly connects the front and the back of the convolution module, which is equivalent to directly transmit the information to the latter part of the network.

In Inception module of GoogLeNet [4], the multiple branches may form a connection similar to the shortcut connection. Since one of the branches has a convolution kernel of 1 and a stride of 1, and the branch is actually a ResNet shortcut connection in the literature [3]. In this way, the Inception module can also reduce the information loss.

In our proposed LSTM system, we connect the cell state flow and the hidden output flow of LSTM. The bi-directional LSTM is used in our system. On one hand, it strengthens the temporal connection of the data, on the other hand, it reduces the loss of information, since the information is connected from many directions.

## 3.4 The structure of the Context-LSTM system

Our experimental system includes a CNN backbone (ResNet-50) and a LSTM classifier (Context-LSTM classifier). This is a common structure of the LSTM system. The backbone is to extract the features of the data, and enter the features into the LSTM system. The structure of the Context-LSTM system is shown in Fig. 1.

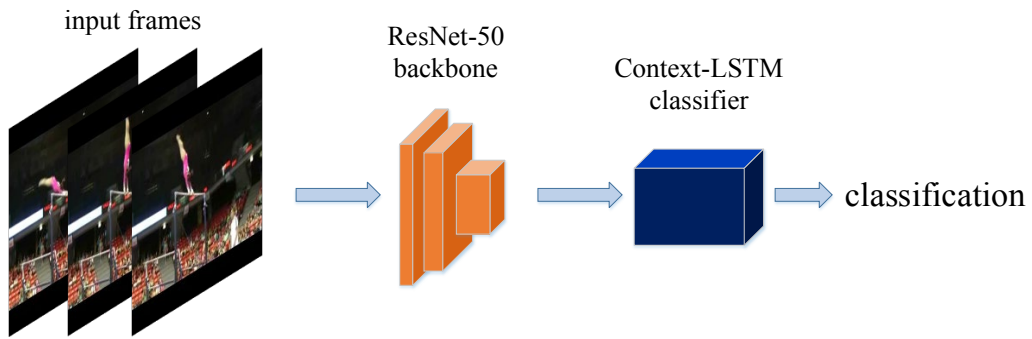

Fig. 1 The structure of the Context-LSTM system. It consists of two parts: a ResNet-50 backbone and a classifier, the classifier is named Context-LSTM classifier. The ResNet-50 backbone is used to extract the shallow and the deep semantic features, the Context-LSTM classifier is used to process the deep temporal features.

LSTM has the ability to extract features either. For video detection, since their features are more complicated, it needs to set a backbone to extract features from the videos. These features are input into LSTM for the temporal feature learning.

The CNN backbone we used is ResNet-50, which was pre-trained by ImageNet [19]. ResNet-50 is not a deep network, and the residual connection further reduces the loss of information. We believe that, ResNet-50 is a network with strong generalization ability, and can extract deep and shallow semantic feature information at the same time.

LSTM is an improvement of RNN. RNN is developed for processing the temporal sequences through an extended neural network chain. LSTM adds the gates into RNN, for solving the problem of gradient disappearance and gradient explosion when RNN processes long temporal sequences.

There are 3 kinds of gates in LSTM, which are forget gate, input gate and output gate. The forget gate determines how much of the cell state $c_{t-1}$ from the previous node will be retained to the current node $c_t$. The input gate determines how much of the input of the network $x_t$ is saved in the cell state $c_t$. The output gate determines how much of the cell state $c_t$ is transmitted to the output value $h_t$.

The operation process of LSTM has three stages. a). Forgetting stage. This stage is to selectively forget the input from the previous node. Namely forget the unimportant and remember the important. b). Selecting the memory stage. This stage selectively memorizes the input of this stage. It is mainly to select and memorize the input $x_t$. c). Output stage. This stage will determine which will be the output of the current node.

The process of LSTM is as follows:

$$f_t = \sigma(W_f \cdot [h_{t-1}, x_t] + b_f) \quad (12)$$
$$i_t = \sigma(W_i \cdot [h_{t-1}, x_t] + b_i) \quad (13)$$
$$\tilde{C}_t = \tanh(W_C \cdot [h_{t-1}, x_t] + b_C) \quad (14)$$
$$C_t = f_t * C_{t-1} + i_t * \tilde{C}_t \quad (15)$$
$$o_t = \sigma(W_o[h_{t-1}, x_t] + b_o) \quad (16)$$
$$h_t = o_t * \tanh(C_t) \quad (17)$$

where $f_t$ is the function of the forget gate at time $t$, $W_f$ is the weight of the forget gate, $h_{t-1}$ is the previous output, $x_t$ is the input, $b_f$ is the bias of the forget gate, $\sigma$ is the sigmoid function, $W_i$ is the weight of the input gate, $b_i$ is the bias of the input gate, $i_t$ is the function of the input gate at time $t$, $\tilde{C}_t$ is the short memory state of the cell at time $t$, $W_C$ is the weight of the cell, $b_i$ is the bias of the cell, $C_t$ is the long memory state of the cell at time $t$, $C_{t-1}$ is the long memory state of the cell at time $t-1$, $o_t$ is the function of the output gate at time $t$, $W_o$ is the weight of the output gate, $b_o$ is the bias of the output gate, $h_t$ is the function of the hidden state at time $t$. Equation 12 reflects the forget gate, Equation 13 denotes the input gate, and Equation 16 indicates the output gate.

The Context-LSTM classifier structure is as follows. The Context-LSTM classifier is designed to learn the deep temporal features. The LSTM system is generated from the LSTM blocks, which is offered by the framework PyTorch [20]. We connect the LSTM blocks layer by layer, and connect the cell state flow and the hidden output flow of each LSTM block. Among the LSTM blocks, we use ReLU as the activation function, Batch Normalization, and MaxPooling function. We use the MaxPooling function here since the tensors of PyTorch in our experiments are always matrixes. The structure of the LSTM classifier is shown in Fig. 2.

## 3.5 The traditional CNN-LSTM system (ResNet-50-LSTM)

Since the traditional CNN-LSTM system was compared in our experiments, we introduce the structure here. The structure of the traditional CNN-LSTM system contains a pre-trained ResNet-50 backbone, which is provided by PyTorch, a LSTM block, which is provided by PyTorch either, and two fully connected (FC) layers. The ResNet-50 backbone is pre-trained by ImageNet. The layers of the LSTM block was set to 3, and the hidden size (number of the units) of the LSTM block was set to 512. The LSTM block is the same as the ones in the propose Context-LSTM.

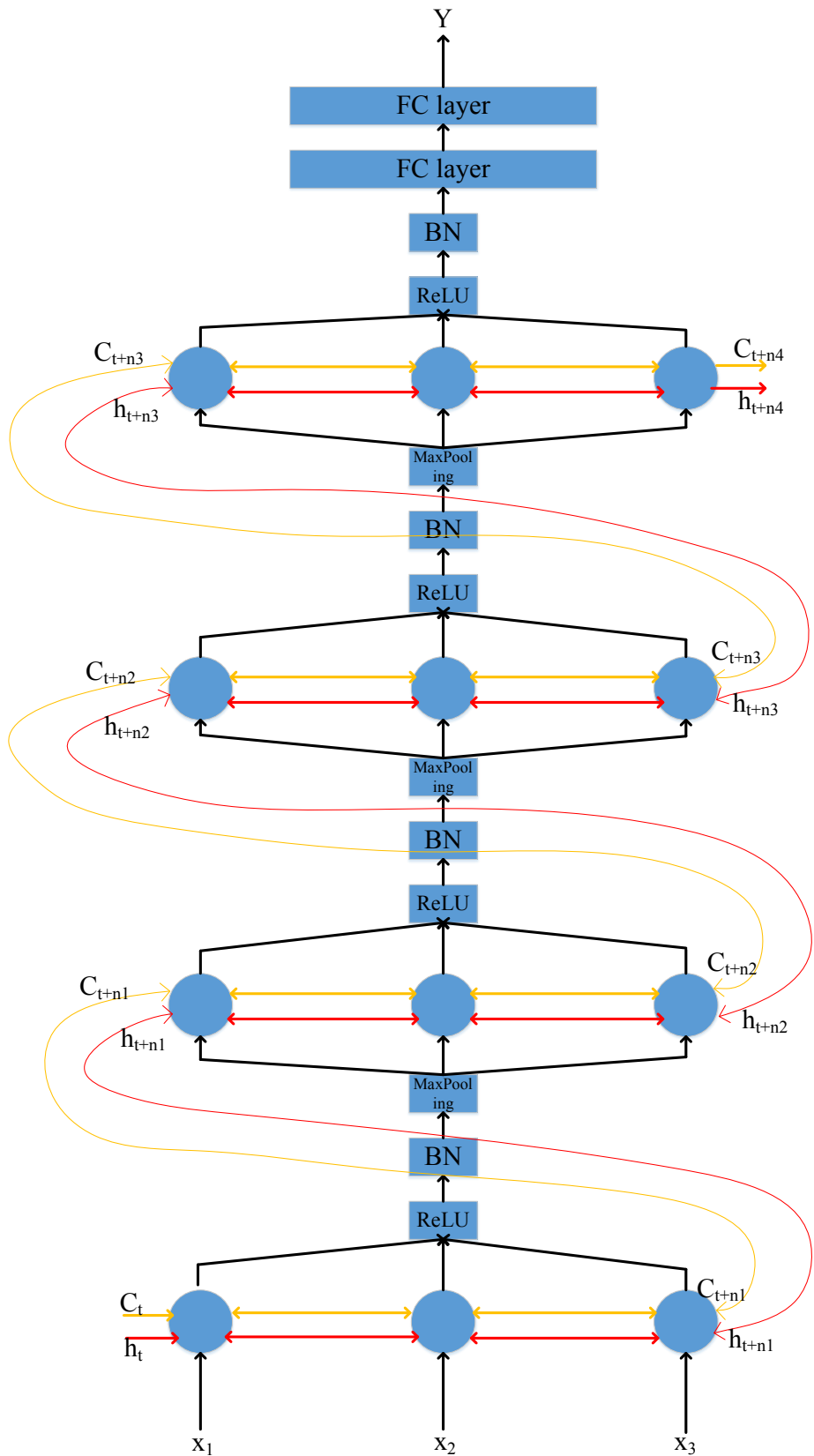

Fig. 2 The structure of the Context-LSTM classifier. Every unit of the LSTM structure has three LSTM layers. $x_1$, $x_2$, $x_3$ is the input data, and Y is the output, which is for the prediction. The LSTM units are the bi-directional LSTM, which is for reducing the information transmission loss.

## 4. Experiments and results

### 4.1 Experiment Setup

The operating system we used was Ubuntu 16.04. The CPU was Intel i7 8700, the RAM was 32GB, the GPU was NVIDIA GeForce 1080ti. PyTorch, Numpy, Sklearn, Matplotlib, Pandas, and tqdm were implemented in the experiments.

The UCF101 dataset was split into frames before the experiment. One video was folded into one folder, and the name of the folder was the annotation of the video. The annotation was extracted during the model training.

### 4.2 Training

We used cross-entropy loss function as the training loss. The detection accuracy function was provided by Sklearn, which was a top-1 accuracy essentially. The training loss and the test accuracy were calculated by the generated model every epoch. The UCF101 dataset was divided into the training set and the test set randomly, by the ratio of 3:1.

The ResNet model was offered by PyTorch, which was a pre-trained model. The dropout function was used in the FC layers and the LSTM blocks. The LSTM block had 3 LSTM layers. The hidden size (number of the units) of LSTM was set to 512. The latter part of the LSTM system was two fully connected (FC) layers for classification. The input and output dimensions of the first FC layer are 512 and 256 respectively, and the input and output dimensions of the second FC layer are 256 and 101 respectively. 101 is the number of categories of the UCF101 dataset.

The epoch was set to 300. The batch size was set to 80. The learning rate was set to 0.001. The algorithm Adam was used as the optimization algorithm. From the experiment, we found that, when increasing the batch size, the test accuracy would be improved. The possible reason is analyzed in the Section Discussion.

### 4.3 Experiment results

In the experiment, we used different classifiers, for the purpose of comparing the proposed Context-LSTM classifier with other classifiers. The other parts of the experiment were all the same. The parameters of the comparison experiments were set all the same, for the purpose of testing the proposed Context-LSTM classifier separately. The Context-LSTM was illustrated in Section 3.5. The single detection visualization results of the Context-LSTM classifier are shown in Fig. 3.

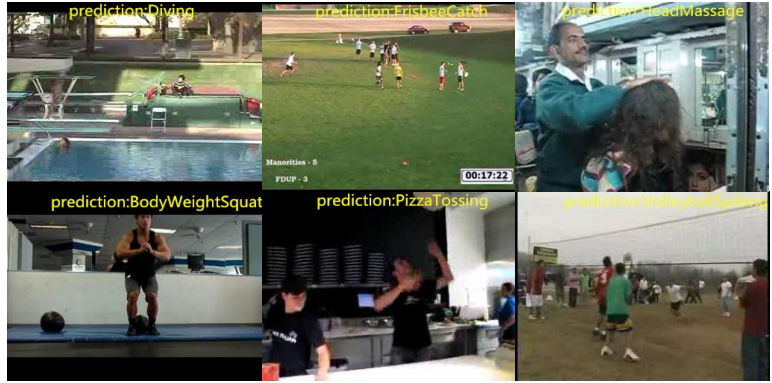

Fig. 3 The visualization detection results of the Context-LSTM classifier.

We compared the proposed classifier with the traditional CNN-LSTM system. The traditional CNN-LSTM system is illustrated in Section 3.6. Compared with the traditional CNN-LSTM system, the proposed Context-LSTM has the following advantages: a) the test accuracy of the proposed Context-LSTM was higher than the traditional CNN-LSTM, which was the state-of-the-art top-1 accuracy on UCF101; b) the training time of the proposed Context-LSTM was much shorter than that of the traditional CNN-LSTM, which was reduced from 93 hours to about 55 hours, under the epoch of 300; c) the GPU memory occupation of the proposed Context-LSTM was smaller than that of the traditional CNN-LSTM, which reduced about 2GB of the GPU memory; d) with the large increasing of epoch, the proposed Context-LSTM did not show the phenomenon of overfitting. It shows that the proposed Context-LSTM has good robustness. The following is a detailed description.

**The test accuracy.** In our experiment, the proposed context LSTM achieved 92.2% top-1 accuracy on UCF101 validation dataset. Our test accuracy was run on the entire validation set, but not only on the selected clips or frames. Fig. 4 is a screen capture of the experiment. Moreover, after the epoch 150, the test accuracy of Context-LSTM was above 91% stably. Fig. 5 is the training loss and the test accuracy of the Context-LSTM classifier on UCF101, where the top-1 validation set accuracy is stable at more than 91% after the epoch 200. In the literatures, other researchers reported a UCF101 accuracy which was not the top-1 accuracy, but a mean value, and the test set (only for computing the accuracy) was some clips. The literature [14] reported that, some other excellent algorithms achieved about 87% top-1 accuracy on 10 clips of UCF101. The literature [13] reported that, the algorithm SMART achieved 74.6% accuracy on the entire validation set. We think our test accuracy may achieve the state-of-the-art top-1 accuracy on the entire validation dataset.

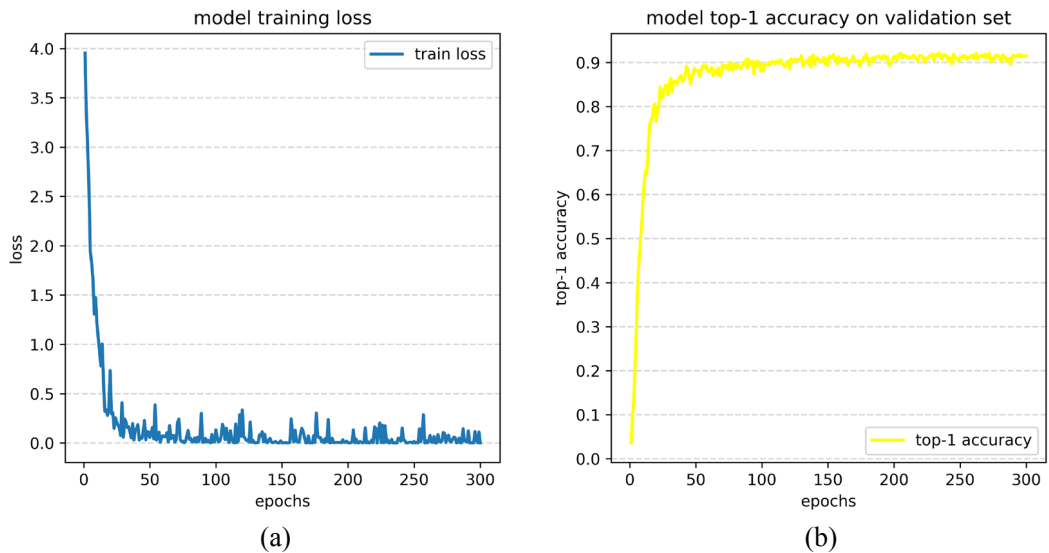

Fig. 4 The screen capture of the Context-LSTM model training. The top-1 accuracy on the validation dataset was computed by the model of every epoch. Moreover, it can be seen from the upper left corner of the screen capture, the current GPU memory usage was 3636MB when the epoch was 235.

Fig. 5 The training loss and the top-1 accuracy on the UCF101 validation dataset of the Context-LSTM classifier. The loss curve is stable and lower, and the accuracy is stable and competitive. The stableness shows the robustness of the Context-LSTM classifier, the model shows a competitive performance on the temporal dataset UCF101.

The traditional CNN-LSTM model reported the test top-1 accuracy at the maximum value of 90% in our experiment, the CNN backbone was ResNet-50 which was pre-trained by ImageNet, the LSTM system was a LSTM block provided by PyTorch. The batch size of the CNN-LSTM model was set to 80 either. The test top-1 accuracy is listed in Table 1, which includes the literature's state-of-the-art algorithm and the CNN-LSTM in our experiment.

**The training time.** From our experiments, the training time of the proposed Context-LSTM classifier was about 55 hours, and the training time of the traditional CNN-LSTM was about 93 hours, under the epoch of 300. We in fact increased the net layers of LSTM, but the training time became shorter on the contrary. We observed that the time of each iteration was reduced indeed. We suppose that the reason may be related to the enhanced utilization of the data temporal information, which is named deep temporal feature information.

**The GPU memory.** Our experiments showed that, compared with the traditional CNN-LSTM, the proposed Context-LSTM spent nearly 2GB less GPU memory. The GPU memory spend of the Context-LSTM was about 3.6 GB, with the batch size of 80 (it is shown in Fig. 4). The GPU memory spend of the traditional CNN-LSTM model was about 5.5 GB, with the same batch size. The reason we analyzed is that, on one hand, the deep temporal information is utilized; on the other hand, the ReLU and MaxPooling operation may reduce the computation.

**The robustness.** From our experiments, Context-LSTM did not overfit. In the experiment with epoch of 300, we observed the loss and accuracy after the 200 epoch. The value of loss was always stable and small, and the accuracy was stably at more than 91% either. These points showed that Context-LSTM had good robustness. Fig. 5 also shows this point.

Table 1. The top-1 accuracy on UCF101(%). ResNet-50-LSTM and ours are the entire validation dataset accuracy, 3D ResNeXt-101 is 10 clips accuracy, SMART is all clips accuracy. ResNet-50-LSTM is the CNN-LSTM in our experiment.

| epoch | 30 | 60 | 90 | 120 | 150 | 180 | 210 | 240 | 270 | 300 | best |
|---|---|---|---|---|---|---|---|---|---|---|---|
| **ResNet-50-LSTM** | 78.7 | 84.5 | 87.8 | 87.9 | 88.6 | 88.5 | 88.7 | 89.2 | 89.5 | 88.8 | 90.7 |
| **3D ResNeXt-101 [14]** | - | - | - | - | - | - | - | - | - | - | 89.8(10 clips) |
| **SMART [13]** | - | - | - | - | - | - | - | - | - | - | 74.6(all clips) |
| **ours** | 85.3 | 87.8 | 90.8 | 90.4 | 89.3 | 90.6 | 91.0 | 91.6 | 91.2 | 91.5 | 92.2 |

## 4.4 The ablation study

We studied the performance of the proposed Context-LSTM classifier by removing the MaxPooling layer, and compared it with the traditional ResNet-50 CNN-LSTM. The other parameters of the experiment were set the same, such as the batch size (80), learning rate (0.001), etc. The top-1 accuracy on the validation set, the training time and the GPU memory occupation are listed in Table 2. The comparison of the traditional ResNet-50 CNN-LSTM and the proposed Context-LSTM classifier is figured in Fig. 6.

From the experiment, it can be seen that, the MaxPooling layer could enhance the test accuracy, accelerate the training time, and reduce the GPU memory usage. In the system, the order of ReLU-BN was better than the order of BN-ReLU, i.e., the training time was faster and the test accuracy was higher. The experimental results showed that, the MaxPooling layer was crucial for the Context-LSTM.

Table 2. The top-1 accuracy on the validation set, the training time and the GPU memory occupation of the models. ResNet-50-LSTM is the traditional CNN-LSTM illustrated in Section 3.5. All other experimental parameters were the same.

|  | accuracy(best)(%) | training time | GPU usage(GB) |
|---|---|---|---|
| **ResNet-50-LSTM** | 90.7 | 93 hours | 5.5 |
| **Context-LSTM without MaxPooling** | 91.0 | 70 hours | 4.5 |
| **Context-LSTM** | **92.2** | **55 hours** | **3.6** |

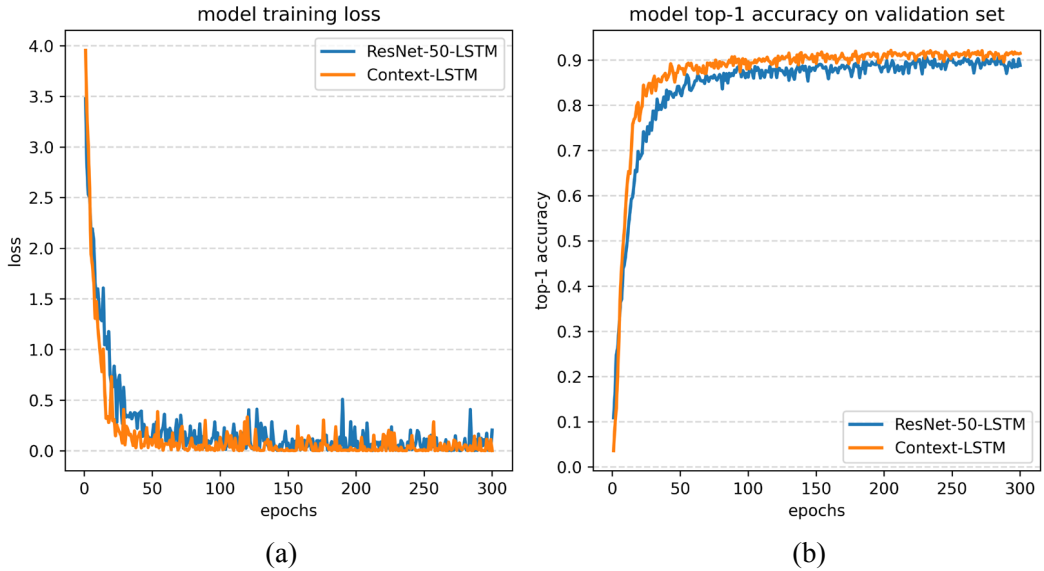

(a) (b)

Fig. 6 The training loss and the top-1 accuracy of the traditional CNN-LSTM (ResNet-50-LSTM) and the Context-LSTM classifier. Compared with the CNN-LSTM, the stableness and robustness of our proposed model are better, the loss of the proposed model converges faster, and there is no large jitter; the detection accuracy of our model rises faster, the accuracy of it is higher and competitive, and the accuracy rate is more stable.

## 5. Discussion

One of the sources of the proposed LSTM system is that, deeper networks are able to extract higher-level semantic features, and high-level semantic features may cause higher detection accuracy. To avoid the information loss caused by the deeper layers, we used the bi-directional LSTM to construct the system. We believe that, the Context-LSTM classifier simulates the cerebral cortex of human.

We notice that, artificial neural networks are similar with biology neural networks in certain aspects. For instance, prior knowledge and transfer learning [21] can improve the detection accuracy of artificial neural network, which can also apply to biological neural networks. The attention mechanism [22] greatly improves the efficiency of machine learning, and improving attention will also improve the efficiency of human learning. We assume that, the LSTM network simulated the neurons of the human brain, since there were forgetting gates, input gates, and output gates in LSTM. In this paper, we simulated an uncomplicated cerebral cortex structure by increasing the number of LSTM layers. And we used the reaction function, BN and pooling function in the Context-LSTM structure to simulate the axons and dendrites in the neurons of the cerebral cortex.

We compare Context-LSTM to human recognition system. Compared with human eyes, the front ResNet-50 backbone can extract complex and deep features; The back LSTM classifier is compared to the cerebral cortex, which can process the deep temporal feature information and classify it. The experimental result is worthy.

# 6. Conclusion

A deep feature structure of LSTM was built to detect deep temporal features, and the Context-LSTM classifier was proposed. On the UCF101 dataset, our proposed structure achieved state-of-the-art results in the top-1 accuracy of the entire validation dataset. The proposed structure could reduce the model training time and the GPU memory usage. Our structure was robust and did not overfit during training.

The Context-LSTM has a simple structure that only has a backbone and a classifier. The Context-LSTM could simulate a human recognition system. The backbone could simulate the human eyes for the reason of extracting different level of features, and the LSTM system could simulate the cerebral cortex for the reason of processing the deep temporal features. The experimental results showed some competitive performance of the proposed system.

# References

[1] Alex Krizhevsky, Ilya Sutskever, and Geoffrey E Hinton, "Imagenet classification with deep convolutional neural networks," in *Advances in neural information processing systems*, 2012, pp. 1097-1105.

[2] Karen Simonyan and Andrew Zisserman, "Very deep convolutional networks for large-scale image recognition," *arXiv preprint arXiv:1409.1556,* 2014.

[3] Kaiming He, Xiangyu Zhang, Shaoqing Ren, and Jian Sun, "Deep residual learning for image recognition," in *Proceedings of the IEEE conference on computer vision and pattern recognition*, 2016, pp. 770-778.

[4] Christian Szegedy, Wei Liu, Yangqing Jia, Pierre Sermanet, Scott Reed, Dragomir Anguelov, Dumitru Erhan, Vincent Vanhoucke, and Andrew Rabinovich, "Going deeper with convolutions," in *Proceedings of the IEEE conference on computer vision and pattern recognition*, 2015, pp. 1-9.

[5] Christian Szegedy, Vincent Vanhoucke, Sergey Ioffe, Jon Shlens, and Zbigniew Wojna, "Rethinking the inception architecture for computer vision," in

| | *Proceedings of the IEEE conference on computer vision and pattern recognition,* 2016, pp. 2818-2826. |
|---|---|
| [6] | C Szegedy, S Ioffe, V Vanhoucke, and A Alemi, "Inception-ResNet and the Impact of Residual Connections on Learning," *arXiv preprint arXiv:1602.07261.* |
| [7] | Jeffrey L Elman, "Finding structure in time," *Cognitive science,* vol. 14, no. 2, pp. 179-211, 1990. |
| [8] | Sepp Hochreiter and Jürgen Schmidhuber, "Long short-term memory," *Neural computation,* vol. 9, no. 8, pp. 1735-1780, 1997. |
| [9] | Xingjian Shi, Zhourong Chen, Hao Wang, Dit-Yan Yeung, Wai-Kin Wong, and Wang-Chun Woo, "Convolutional LSTM network: A machine learning approach for precipitation nowcasting," *Advances in neural information processing systems,* vol. 28, 2015. |
| [10] | Haodong Duan, Yue Zhao, Yuanjun Xiong, Wentao Liu, and Dahua Lin, "Omni-Sourced Webly-Supervised Learning for Video Recognition," Cham, 2020: Springer International Publishing, in Computer Vision – ECCV 2020, pp. 670-688. |
| [11] | Xianyuan Wang, Zhenjiang Miao, Ruyi Zhang, and Shanshan Hao, "I3D-LSTM: A New Model for Human Action Recognition," *IOP Conference Series: Materials Science and Engineering,* vol. 569, no. 3, p. 032035, 2019/07/01 2019, doi: 10.1088/1757-899x/569/3/032035. |
| [12] | Guoxi Huang and Adrian G Bors, "Busy-Quiet Video Disentangling for Video Classification," in *Proceedings of the IEEE/CVF Winter Conference on Applications of Computer Vision,* 2022, pp. 1341-1350. |
| [13] | Shreyank N Gowda, Marcus Rohrbach, and Laura Sevilla-Lara, "SMART Frame Selection for Action Recognition," in *Proceedings of the AAAI Conference on Artificial Intelligence,* 2021, vol. 35, no. 2, pp. 1451-1459. |
| [14] | Shervin Manzuri Shalmani, Fei Chiang, and Rong Zheng, "Efficient Action Recognition Using Confidence Distillation," *arXiv preprint arXiv:2109.02137,* 2021. |
| [15] | Chunting Zhou, Chonglin Sun, Zhiyuan Liu, and Francis Lau, "A C-LSTM neural network for text classification," *arXiv preprint arXiv:1511.08630,* 2015. |
| [16] | Sergey Ioffe and Christian Szegedy, "Batch normalization: Accelerating deep network training by reducing internal covariate shift," in *International conference on machine learning,* 2015: PMLR, pp. 448-456. |
| [17] | Khurram Soomro, Amir Roshan Zamir, and Mubarak Shah, "UCF101: A dataset of 101 human actions classes from videos in the wild," *arXiv preprint arXiv:1212.0402,* 2012. |
| [18] | Alexey Dosovitskiy, Philipp Fischer, Eddy Ilg, Philip Hausser, Caner Hazirbas, Vladimir Golkov, Patrick Van Der Smagt, Daniel Cremers, and Thomas Brox, "Flownet: Learning optical flow with convolutional networks," in *Proceedings of the IEEE international conference on computer vision,* 2015, pp. 2758-2766. |
| [19] | Jia Deng, Wei Dong, Richard Socher, Li-Jia Li, Kai Li, and Li Fei-Fei, "Imagenet: A large-scale hierarchical image database," in *2009 IEEE conference* |

*on computer vision and pattern recognition*, 2009: Ieee, pp. 248-255.

[20] Adam Paszke, Sam Gross, Francisco Massa, Adam Lerer, James Bradbury, Gregory Chanan, Trevor Killeen, Zeming Lin, Natalia Gimelshein, and Luca Antiga, "PyTorch: An Imperative Style, High-Performance Deep Learning Library," *Advances in Neural Information Processing Systems,* vol. 32, pp. 8026-8037, 2019.

[21] Ly Pratt and Sebastian Thrun, "Machine Learning-Special Issue on Inductive Transfer," 1997.

[22] Ashish Vaswani, Noam Shazeer, Niki Parmar, Jakob Uszkoreit, Llion Jones, Aidan N Gomez, Łukasz Kaiser, and Illia Polosukhin, "Attention is all you need," in *Advances in neural information processing systems*, 2017, pp. 5998-6008.